\documentclass[conference,9pt]{IEEEtran}
\usepackage{array}
\usepackage{cite}
\usepackage{amsmath,amssymb,amsfonts}
\usepackage{booktabs}
\usepackage{colortbl}
\usepackage{enumitem}
\usepackage{fvextra}
\usepackage{graphicx}
\usepackage{hyperref}
\usepackage{ifthen}
\usepackage{multirow}
\usepackage{stfloats}
\usepackage{subcaption}
\usepackage{xcolor}

\newcommand{\N}{\mathbb{N}}
\newcommand{\R}{\mathbb{R}}
\newcommand{\graco}{\texorpdfstring{$GraCo$}{GraCo}}
\newcommand{\ngspice}{ngspice}

\newboolean{anonymous}
\setboolean{anonymous}{false} 

\usepackage[acronym]{glossaries}
\newacronym{eda}{EDA}{electronic design automation}
\newacronym{llm}{LLM}{large language model}
\newacronym{ppa}{PPA}{power, performance, area}
\newacronym{dut}{DUT}{device under test}
\newacronym{inv}{INV}{inverter}
\newacronym{nand2}{NAND2}{two-input NAND gate}
\newacronym{ff}{FF}{flip-flop}
\newacronym{pdk}{PDK}{process design kit}
\newacronym{ip}{IP}{intellectual property}
\glsdisablehyper

\colorlet{tablerowcolor}{gray!10} 
\newcommand{\rowcol}{\rowcolor{tablerowcolor}} %

\begin{document}

\title{\graco{} -- A Graph Composer for Integrated Circuits}

\ifthenelse{\boolean{anonymous}}{
    \author{
        \IEEEauthorblockN{Anonymous Authors}
        \IEEEauthorblockA{Anonymous Affiliations}}
}{
    \author{
        \IEEEauthorblockN{Stefan Uhlich\textsuperscript{1}, Andrea Bonetti\textsuperscript{2,4}, Arun Venkitaraman\textsuperscript{2,4}, Ali Momeni\textsuperscript{2,3}}
        \IEEEauthorblockN{Ryoga Matsuo\textsuperscript{1,3}, Chia-Yu Hsieh\textsuperscript{2,4}, Eisaku Ohbuchi\textsuperscript{4}, Lorenzo Servadei\textsuperscript{2,4}}
        \IEEEauthorblockA{\textsuperscript{1}\textit{Sony Semiconductor Solutions Europe, Germany}\quad \textsuperscript{2}\textit{SonyAI, Switzerland}\quad \textsuperscript{3}\textit{EPFL, Switzerland}\quad \textsuperscript{4}\textit{Sony Semiconductor Solutions, Japan}}}
}


\maketitle

\begin{abstract}
Designing integrated circuits involves substantial complexity, posing challenges in revealing its potential applications --- from custom digital cells to analog circuits.
Despite extensive research over the past decades in building versatile and automated frameworks, there remains open room to explore more computationally efficient AI-based solutions. This paper introduces the graph composer \graco{}, a novel method for synthesizing integrated circuits using reinforcement learning~(RL). \graco{} learns to construct a graph step-by-step, which is then converted into a netlist and simulated with SPICE. We demonstrate that \graco{} is highly configurable, enabling the incorporation of prior design knowledge into the framework. We formalize how this prior knowledge can be utilized and, in particular, show that applying consistency checks enhances the efficiency of the sampling process. To evaluate its performance, we compare \graco{} to a random baseline, which is known to perform well for smaller design space problems. We demonstrate that \graco{} can discover circuits for tasks such as generating standard cells, including the inverter and the two-input NAND (NAND2) gate. Compared to a random baseline, \graco{} requires $5\times$ fewer sampling steps to design an inverter and successfully synthesizes a NAND2 gate that is $2.5\times$ faster.

\end{abstract}


\section{Introduction}
\label{sec:intro}

\begin{figure*}[!b]
    \vspace*{-0.3cm}
    \resizebox{\linewidth}{!}{\includegraphics[trim=4 5 4 5,clip]{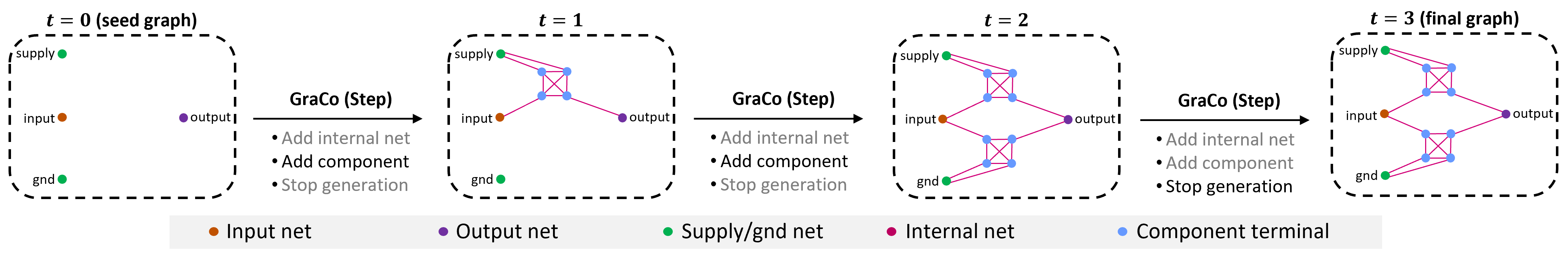}}
    \caption{Auto-regressive graph generation in \graco{}. At each step, it decides whether to add an internal net, insert a component, or stop the generation process. As an example, we illustrate the design of an inverter consisting of an NMOS and a PMOS transistor.}
    \label{fig:sequential_generation}
\end{figure*}


\begin{figure*}
    \resizebox{\linewidth}{!}{\includegraphics[trim=3 3 0 5,clip]{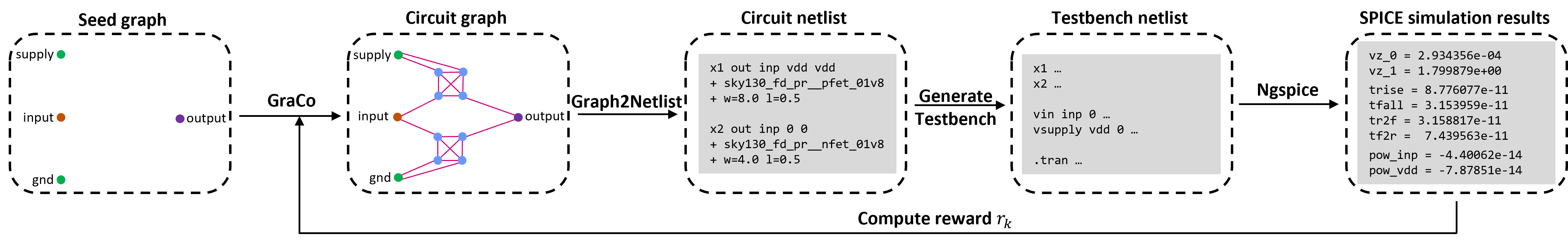}}
    \caption{Overall framework of \graco{}. As an example, we show the design of an inverter with a NMOS and PMOS transistor.}
    \label{fig:overall_framework}
    \vspace{-0.4cm}
\end{figure*}

\noindent Designing integrated circuits is a challenging and time-intensive process, even for experienced design engineers. Consequently, there is increasing interest in efficient and automated circuit design methods to reduce labor-intensive efforts, enhance productivity, and improve design accuracy~\cite{sripramong2002invention,mitea2011topology,meissner2014feats,yang2017smart,rojec2019analog,huss2006analog,settaluri2020autockt,zhao2020automated,fan2021specification,budak2021dnn,zhao2022analog,chen2023total,lu2023automatic,dong2023cktgnn,budak2023apostle,zhao2024automated}. Circuit synthesis can be broadly divided into two main steps: \emph{(i) discovering the circuit topology}, and \emph{(ii) sizing the circuit parameters}, which may be closely interdependent or performed sequentially.

Previous works, such as~\cite{yang2017smart,settaluri2020autockt,budak2021dnn,budak2023apostle}, have achieved notable results in circuit sizing but are restricted to fixed circuit topologies. This limitation arises because topology search is significantly more complex due to the exponential growth of the design space with increasing circuit size. For example, the complexity of circuit synthesis is underscored by the fact that the number of possible wiring configurations is determined by the Bell number $B_n$~\cite{aigner1999characterization,graeb2001sizing} which is quickly growing. For instance, with two MOSFETs one can already generate more than fifty useful circuits~\cite{pretl2021fifty}. When considering that an inventory often includes multiple component types and that each selected component requires sizing, the complexity of the problem becomes even more evident.

This paper proposes a versatile RL framework for circuit synthesis that integrates both topology exploration and sizing optimization. Leveraging the graph representation of circuits, we introduce \textbf{gra}ph \textbf{co}mposer, named \graco{}, where graphs are constructed autoregressively and transformed into netlists for final circuit simulation in SPICE. To improve sampling efficiency, we incorporate novel consistency checks that reduce complexity and enhance \graco{}'s performance. Furthermore, we show that the framework can seamlessly integrate design knowledge, such as rules required for manufacturing constraints. We evaluate our approach on two circuit synthesis tasks: an inverter gate and a two-input NAND gate~(NAND2).

Previous research has often relied on a fixed library of basic building blocks with structural wiring rules that define how these components can be combined to create a circuit~\cite{sripramong2002invention,mitea2011topology,meissner2014feats,huss2006analog,zhao2020automated,zhao2022analog,zhao2024automated}. Additionally, much of this work has focused on specific applications, such as designing operational amplifiers (OpAmps)~\cite{sripramong2002invention,lu2023automatic,dong2023cktgnn} or power converters~\cite{fan2021specification}, limiting its generality and often assuming the existence of effective sizing methods~\cite{chen2023total} or even using fixed sizing for each component~\cite{fan2021specification}. In contrast, \graco{} overcomes these limitations. It can work with any component, e.g., directly with transistors, without requiring predefined structural wiring rules, offering greater flexibility. At the same time, it can incorporate predefined building blocks using subcircuits if needed. Moreover, the \graco{} framework is versatile, capable of addressing a wide variety of tasks due to its straightforward task definition process and relying on SPICE simulations to verify performance, as demonstrated later, making it suitable for diverse applications. Finally, \graco{} learns both the topology and the sizing of components simultaneously, addressing a key limitation of existing sizing methods, which often focus on a single application, such as OpAmp sizing.

\noindent The key contributions of this work are as follows:
\begin{itemize}
    \item We propose a versatile and autoregressive RL framework called \graco{} for circuit synthesis. Its versatility lies in its ability to integrate diverse design knowledge and handle any circuit elements, including devices (e.g., transistors) and subcircuits. The autoregressive nature allows \graco{} to generate circuits of arbitrary size and autonomously determine when to stop and initiate the SPICE evaluation.
    \item We leverage the analogy between graphs and circuits to enhance circuit synthesis. Graph consistency checks are used to eliminate invalid circuits that need not be simulated in SPICE, improving sampling efficiency.
    \item We provide a detailed discussion on how to effectively set up design tasks and incorporate expert design knowledge in multiple ways to tackle versatile synthesis challenges.
\end{itemize}

\section{Circuit Design with \graco{}}
\label{sec:graco}

\noindent In this section, we will explain \graco{}, our proposed graph composing approach for the design of integrated circuits.

\subsection{Graph Representation of Circuits}
\label{sec:graco:subsec:representation}

\noindent It is well-established that circuits can be represented as graphs, as shown in studies such as~\cite{ohlrich1993subgemini,kunal2020gana,ren2020paragraph,hakhamaneshi2022pretraining}. In this paper, we use a \emph{nets and terminals as nodes} representation. This means that each net and each terminal of a component are represented as graph nodes, with terminals of the same component forming a complete subgraph. This graph representation does not require edge features to distinguish between different terminals of a component, like in a MOSFET.

We use the following features for each node: The first feature identifies the node type (either a net or a terminal). For net nodes, the second feature specifies the net type, which can be ground, internal, input, output, or supply, and the third feature is an index (used only for input, output, or supply to distinguish between multiple inputs, for example). For terminal nodes, the second feature is the terminal index and the remaining features describe the component type and its parameters, e.g., transistor dimensions.

\subsection{Auto-regressive Graph Assembling using \graco{}}
\label{sec:graco:subsec:assembling}

\noindent Using the analogy between circuits and graphs, the problem of circuit design is equivalent to generating a graph. Hence, we propose an auto-regressive approach, where in each step, \graco{} bases its decision which nodes and edges to add on the current partially assembled graph. At each step, a single action can be taken from the following list: \emph{(a) add internal net}, i.e., add one net node to the graph, \emph{(b) add component}, i.e., add terminal nodes forming a complete subgraph and connect each of them with an edge to one net node, and, \emph{(c) stop generation} and continue with SPICE evaluation. This sequential generation process is shown in Fig.~\ref{fig:sequential_generation}.

More explicitly, let $\mathbf{x}_t \in \R^{n_{nt} \times f}$ and $\mathbf{e}_t \in \N^{2 \times n_{et}}$ represent the node feature matrix and the edge connection matrix, respectively, with $n_{nt}$ and $n_{et}$ denoting the number of nodes and edges at time index $t$, and $f$ the number of features. At each step $t$, we compute
\begin{equation}
    \{\mathbf{a}_t, \mathbf{c}_t, \boldsymbol{\mu}_t, \log(\boldsymbol{\sigma}_t),\mathbf{T}_t\} = \text{\graco{}}\left(\mathbf{x}_t, \mathbf{e}_t\right),
    \label{eq:graco}
\end{equation}
where $\mathbf{a}_t \in \R^3$ is a vector of logits for a categorical distribution used to decide which of the three possible actions to take. If we choose to add a new component (see action \emph{(b)} above), we use the logits in $\mathbf{c}_t \in \R^{n_c}$ to decide which of the $n_c$ components to add and $\boldsymbol{\mu}_t \in \R^{n_p}$ and $\log(\boldsymbol{\sigma}_t) \in \R^{n_p}$ as the mean and log-standard deviation of a Gaussian distribution from which we sample the $n_p$ parameters of the component\footnote{For convenience, we have only shown one $\boldsymbol{\mu}_t$ and $\log(\boldsymbol{\sigma}_t)$, although there are $n_c$ such pairs — one for each component, each potentially with a different number of parameters $n_p$.}, respectively. Finally, $\mathbf{T}_t \in \R^{n_{st} \times n_v}$ is the matrix of logits for another categorical distribution, which is used to sample, for each of the $n_v$ terminals, the nets to which it should be connected from the $n_{st}$ net nodes.

\graco{} employs a DeeperGCN architecture~\cite{li2019deepgcns,li2020deepergcn}, an improved version of \emph{graph convolutional network}s (GCNs)~\cite{zhang2019graph} including skip connections. We use eight GCN layers after which we perform mean pooling on the node embeddings. This pooled embedding is then passed through linear layers to produce $\mathbf{a}_t, \mathbf{c}_t, \boldsymbol{\mu}_t$ and $ \log(\boldsymbol{\sigma_t})$. Eventually, $\mathbf{T}_t$ is computed as the inner product between the node features of net nodes and learned embeddings for the terminals.

\subsection{Reinforcement Learning of \graco{}}
\label{sec:graco:subsec:rl}

\noindent To use \graco{}, the designer has to define a \emph{task} consisting of \emph{(i) the number of input, supply, and output nets}, \emph{(ii) a list of available components along with their allowable size ranges}, \emph{(iii) testbench(es)} for evaluating the generated circuit’s performance, and \emph{(iv) a reward computation function} that converts SPICE simulation results into a reward for \graco{}'s reinforcement learning training.

Fig.~\ref{fig:overall_framework} illustrates the training framework we employ. After sampling the graph, a graph-to-netlist conversion is performed to generate the netlist for the circuit. This netlist is then integrated into the testbench(es) and simulated with SPICE. Based on the simulation results, we calculate the reward, which is used to update the weights $\boldsymbol{\theta}$ of \graco{}. We utilize two different RL approaches:

Our first approach is \emph{REINFORCE with a leave-one-out baseline} (RLOO)~\cite{williams1992simple,kool2019buy}, where we minimize the loss function $\mathcal{L}$ with
\begin{equation}
    \mathcal{L}(\boldsymbol{\theta}) = - \sum_{k=0}^{K-1} \sum_{t=0}^{T-1} \frac{r_{k} - \hat{\mu}_{\neg k}}{\hat{\sigma}_{\neg k}} \log \pi_{\boldsymbol{\theta}}(\mathbf{a}_t \vert \mathbf{s}_t) + \lambda \mathcal{H}(\pi_{\boldsymbol{\theta}}(\cdot \vert \mathbf{s}_t)),
\end{equation}
where $\mathbf{a}_t$ represents the action taken by \graco{} at step $t$ in state $\mathbf{s}_t = \{\mathbf{x}_t,\mathbf{e}_t\}$, corresponding to the graph assembled up to that time. Here, $K$ is the number of samples in a minibatch, and $\hat{\mu}_{\neg k}=\sum_{l\ne k} r_k$, $\hat{\sigma}^2_{\neg k} = \sum_{l\ne k} (r_k - \hat{\mu}_{\neg k})^2$ are the leave-one-out mean and variance to locally baseline the rewards, respectively. $\lambda$ controls the strength of the entropy regularization term $\mathcal{H}$.

As a second approach, we apply \emph{evolution strategies}~(ES) as in~\cite{salimans2017evolution}. Here, we perturb the weights $\boldsymbol{\theta}$ with Gaussian noise $\boldsymbol{\epsilon}_k\sim\mathcal{N}(\mathbf{0}, \sigma_\text{ES}^2\mathbf{I})$ before obtaining the $K$ samples in a minibatch and use the weight gradient $\boldsymbol{\nabla}_{\boldsymbol{\theta}}$ with
\begin{equation}
    \boldsymbol{\nabla}_{\boldsymbol{\theta}} = \frac{1}{K\sigma_\text{ES}^2}\sum\nolimits_{k=0}^{K-1} \frac{r_{k} - \hat{\mu}_{\neg k}}{\hat{\sigma}_{\neg k}} \boldsymbol{\epsilon}_k
    \label{eq:es_grad}
\end{equation}
to update the weights in \graco{}. Compared to RLOO, ES offers more exploration, which is advantageous for our synthesis task. Following~\cite{salimans2017evolution}, we implement mirrored sampling~\cite{geweke1988antithetic}, which randomly samples the first $K/2$ perturbations and sets $\boldsymbol{\epsilon}_k = -\boldsymbol{\epsilon}_{k-K/2}$ to generate the samples for the second half. This approach helps to reduce the variance of the ES gradient in \eqref{eq:es_grad}. We will compare both approaches, RLOO and ES, in Sec.~\ref{sec:results}.

\subsection{Discussion of \graco{}}
\label{sec:graco:subsec:discussion}

\noindent In the following, we discuss our proposed \graco{}. Firstly, it is important to note that the synthesis task formulated in the previous section can also be viewed as a planning problem, where \graco{} seeks to determine the optimal assembly strategy to construct the graph that maximizes the reward. Both, RL and planning are conceptually similar problems and share many approaches. As we will demonstrate later, \graco{} can be extended by incorporating additional inputs that provide information about the current (partially) assembled graph. This transitions the problem into a more classical RL setup, where an agent takes actions in response to an environment with inherent uncertainty.

Secondly, defining the task description for the circuit designer is straightforward, as it involves specifying the elements outlined in Sec.~\ref{sec:graco:subsec:rl}. However, it is essential to design an effective reward signal since it will directly influence the circuits \graco{} discovers. Our experiments show that the reward should guide the search, e.g., by using a squared error loss to the expected output voltage, allowing \graco{} to gradually approach the target through ``hill climbing'' as opposed to finding a ``needle-in-a-haystack'' circuit. Additionally, an incentive reward for achieving a target value (e.g., a specific voltage level) can help \graco{} maintain a good solution rather than sacrificing it for other potential improvements. This approach was applied in the examples in Sec.~\ref{sec:results}.

Finally, it is worth noting that we are only interested in identifying the single best instance rather than the final learned policy, which distinguishes this from typical RL settings. Thus, exploration should in general be prioritized over exploitation, and the training, as described in Sec.~\ref{sec:graco:subsec:rl}, serves primarily to direct the search toward promising regions in the design space.

\section{Utilizing Design Expert Knowledge}
\label{sec:expert_knowledge}

\begin{figure}
    \centering
    \includegraphics[width=0.24\linewidth,trim=11 5 12 5,clip]{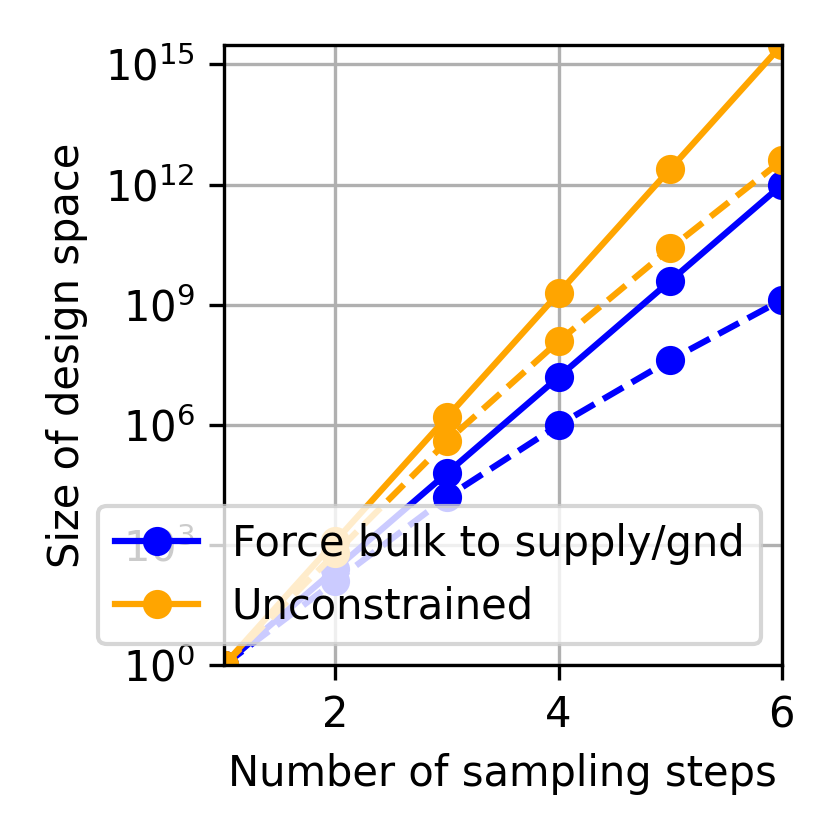}
    \includegraphics[width=0.24\linewidth,trim=11 5 12 5,clip]{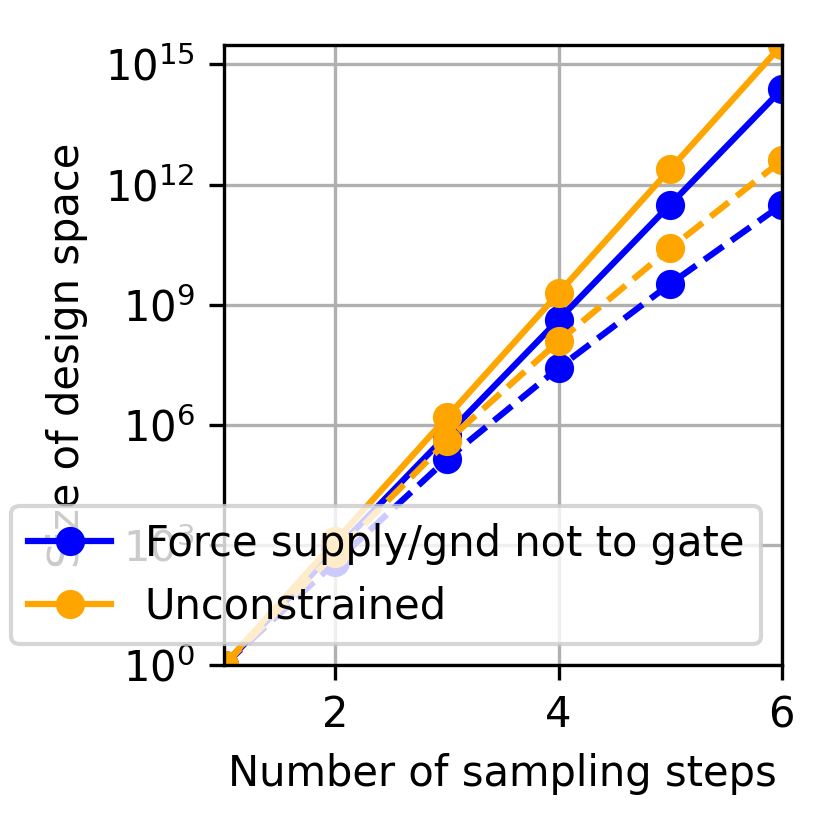}
    \includegraphics[width=0.24\linewidth,trim=11 5 12 5,clip]{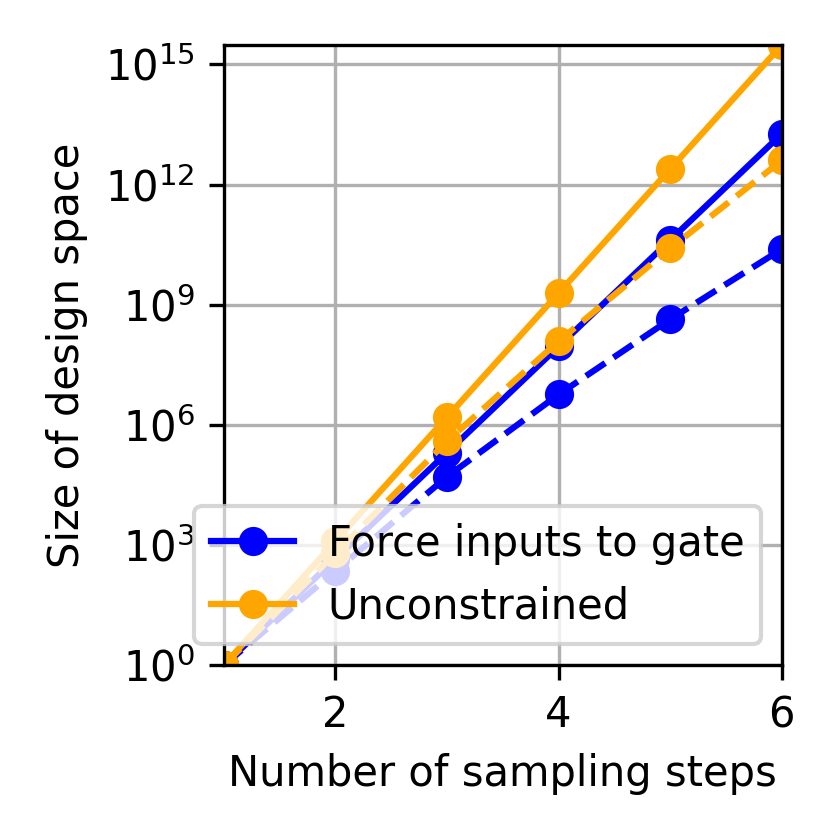}
    \includegraphics[width=0.24\linewidth,trim=11 5 12 5,clip]{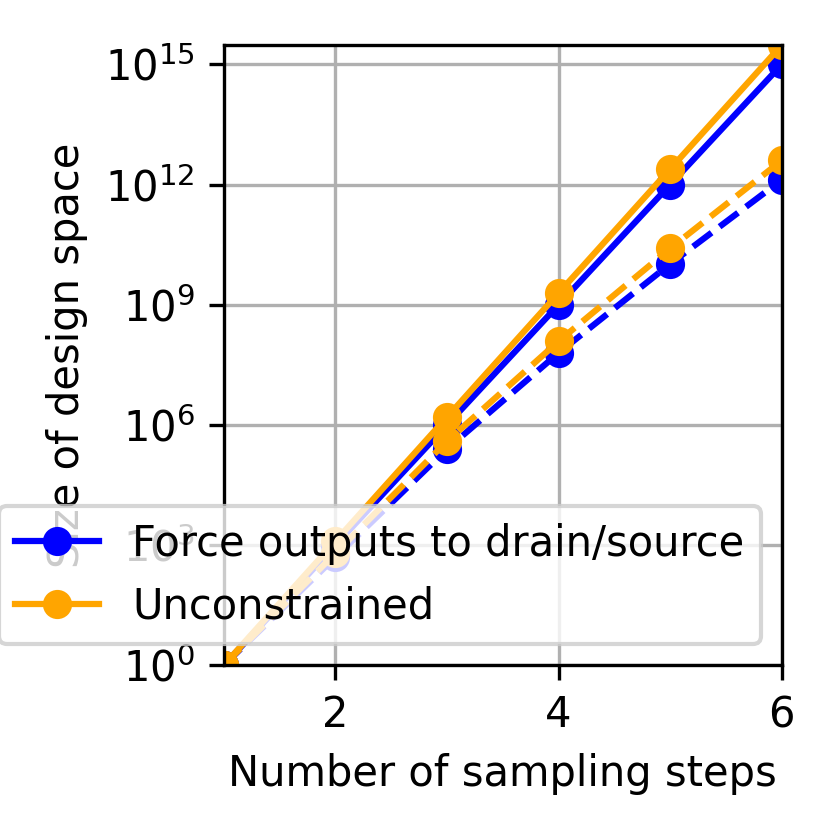}
    \caption{Bounds on NAND2 topology design space size (solid: upper bound, dashed: lower bound). Orange curves show the original design space, while blue curves reflect the effect of explicit wiring rules.}
    \label{fig:search_space}
    \vspace{-0.2cm}
\end{figure}
\begin{figure}
    \centering
    \includegraphics[width=0.49\linewidth,trim=5 5 0 5,clip]{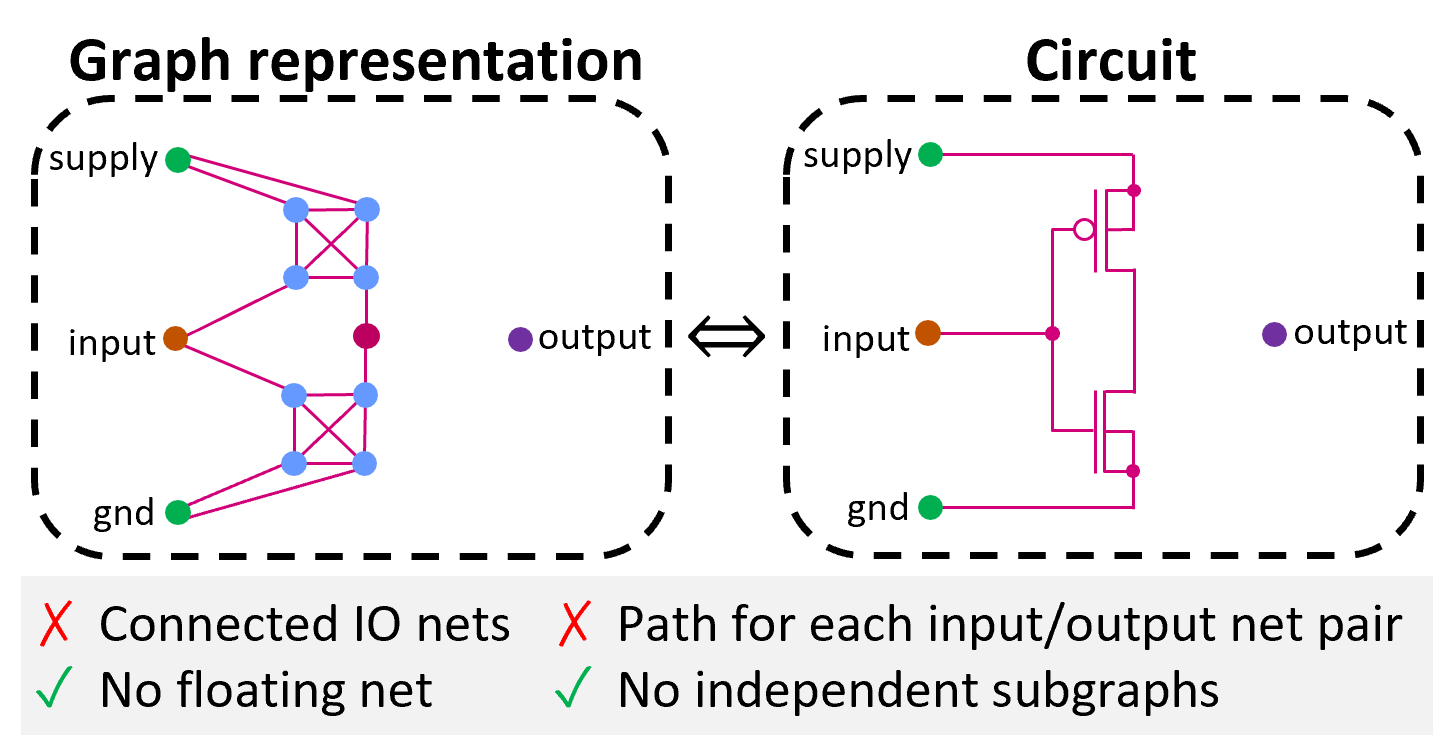}
    \hfill
    \includegraphics[width=0.49\linewidth,trim=5 5 0 5,clip]{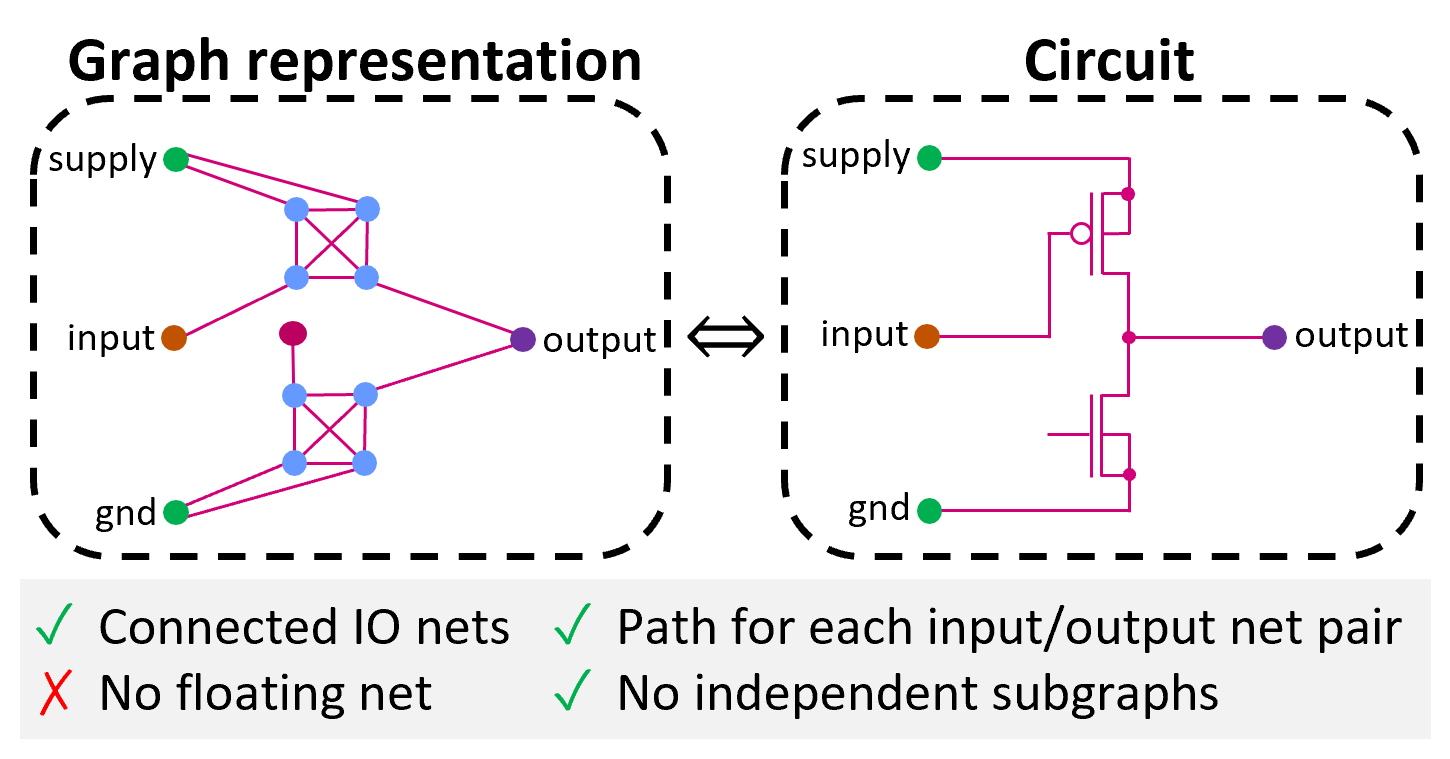}
    \caption{Two examples of consistency checks. These checks are used to ensure that the sampled graphs represent valid circuits.}
    \label{fig:consistency_checks}
    \vspace{-0.3cm}
\end{figure}

\noindent Incorporating design knowledge is essential, as observed in~\cite{meissner2014feats}, to reduce the design space's size. For example, consider designing a NAND gate with two inputs and one output net. Allowing \graco{} to synthesize this gate using NMOS and PMOS transistors requires a total of 6 steps (2 NMOS, 2 PMOS, 1 internal net, and a stop action) to produce the ``standard'' solution. From Fig.~\ref{fig:search_space}, we see that the design space includes between $10^{12}$ and $10^{15}$ possible circuit configurations\footnote{In Fig.~\ref{fig:search_space}, the upper bound assumes all sequence permutations to be unique, ignoring isomorphisms, while the lower bound assumes that an arbitrary step order will result in the same circuit and equivalence among source/drain terminals. The true design space size lies between these bounds since some sequences are isomorphic but also not all step permutations are valid (e.g., some components depend on the presence of an internal net).}. Therefore, reducing or guiding the design space search is crucial for finding good circuits.

There are several methods to introduce additional domain knowledge into \graco{}:
\begin{itemize}
    \item \emph{Reward shaping} to encourage desired circuit designs (e.g., by assigning input nets only to MOSFET gates). This can be implemented by adding a positive reward when the rule is followed or a negative reward when it is not.
    \item \emph{Setting min/max component and internal net limits} if prior knowledge exists about the desired circuit. These constraints can be enforced by adjusting logits: if the minimum number of components/internal nets has not been met, the logit for the \emph{stop generation} action is set to $-\infty$, preventing premature stopping. Similarly, if the maximum number has been reached, the logit for the relevant action is set to $-\infty$, ensuring circuits stay within the specified range.
    \item \emph{Subcircuits} that \graco{} can reuse as predefined IP blocks in the inventory besides elementary blocks, e.g., transistors.
    \item \emph{Graph consistency checks} during sampling to avoid stopping too early (e.g., ensuring all input and output nets are connected). These checks may also involve restarting graph generation if the final graph results in an invalid circuit. We will explore this further in the next subsection.
    \item \emph{Explicit wiring rules} during generation, such as assigning the bulk terminal to supply for PMOS or ground for NMOS, respectively.
\end{itemize}
We will now discuss \emph{consistency checks} and \emph{explicit wiring rules} in detail. The other methods will be revisited in Sec.~\ref{sec:results}. Please note that none of the methods presented here are applied by default, except for the explicit wiring rules that ensure manufacturability, as discussed in Sec.~\ref{sec:expert_knowledge:subsec:hard_constraints}.

\subsection{Graph Consistency Checks}
\label{sec:expert_knowledge:subsec:checks}

\noindent The idea of the consistency checks is to exploit the relationship between the assembled graph and the underlying circuit as discussed in Sec.~\ref{sec:graco:subsec:representation}. Using them, we can ensure that the graph corresponds to a valid circuit, one that can be properly evaluated with SPICE. These checks do not depend on the specific components or tasks, making them applicable in general. We implemented the following consistency checks:
\begin{itemize}
    \item \emph{Connected input/output nets}: Ensures that every input and output net node is connected to at least one terminal node, confirming their link to a component.
    \item \emph{Path between input and output nets}: Verifies that there is a path through the graph connecting each input net node to every output net node.
    \item \emph{No floating nets}: Ensures that all internal or output nodes are either unconnected or connected to at least two terminal nodes, preventing floating nodes during SPICE simulation. Additionally, it ensures that the circuit is connected to at least two different input or supply nets or to ground.
    \item \emph{No isolated subgraphs}: Ensures that the graph does not contain any isolated subgraphs to avoid unnecessary circuits and reduce the risk of potential SPICE convergence errors.
\end{itemize}
Fig.~\ref{fig:consistency_checks} shows two examples for these consistency checks. They can be applied in three ways:
\begin{itemize}
    \item \emph{During generation}: Set the logit for the stop generation action to $-\infty$ if a consistency check fails.
    \item \emph{After generation}: Discard the generated circuit if it fails a consistency check and ask \graco{} to generate a new one. This is done for a maximum number of trials set by the user.
    \item \emph{As additional \graco{} input}: Provide the GCN with information on consistency checks for the current graph, enabling it to take appropriate actions.
\end{itemize}
Furthermore, they can be combined, and each can consider a different subset of consistency checks, offering flexibility in setting up the synthesis task. \graco{} supports this customization, allowing users significant control over the synthesis process.

\subsection{Explicit Wiring Rules}
\label{sec:expert_knowledge:subsec:hard_constraints}
\noindent When launching \graco{}, additional explicit wiring rules can be specified. The following list outlines rules that are particularly useful for designing digital standard cells, as studied in Sec.~\ref{sec:results}:
\begin{itemize}
    \item \emph{Force bulk to supply/gnd}: This rule connects the bulk to ground or supply for NMOS or PMOS, respectively. It is mandatory for standard cells to avoid an unacceptable silicon area penalty caused by the need for dedicated deep wells for each standard cell and here they are considered as a manufacturing constraint given their relevance.
    \item \emph{Force supply/gnd not to gate}: This rule prohibits connecting supply or ground to the gate of a transistor. It is mandatory to prevent electrostatic discharge (ESD) violations.
    \item \emph{Force inputs to gate}: This rule might be requested depending on specifications (e.g., input with high impedance).
    \item \emph{Force outputs to drain/source}: This rule is also typically preferred (e.g., output with low impedance).
\end{itemize}
For standard cells, such as those considered later in this paper, the first two rules are necessary to pass manufacturing checks and are therefore enabled by default in \graco{}. The last two rules are optional and can be enabled by the user as required. Please note that this list is not exhaustive, and additional rules can be added to \graco{} depending on the application.

To analyze the impact of these explicit wiring rules on the size of the design space, we revisit the NAND2 gate design case discussed at the beginning of Sec.~\ref{sec:expert_knowledge}. Fig.~\ref{fig:search_space} illustrates how these rules reduce the size of the design space for the circuit topology. Notably, the \emph{force bulk to supply/gnd} rule significantly decreases the size, as it imposes a fixed assignment of the bulk to a net. Nevertheless, the overall size of the search space remains huge as can also be seen later in Table~\ref{tab:tasks}.

\section{Considered Design Tasks}
\label{sec:tasks}

\noindent In this section, we outline the key decisions involved in defining the design tasks. While the specific choices applied in the experiments presented in this work are reported, these decisions can always be adjusted to meet the user’s requirements.
The decisions include the selection of the targeted circuit and its verification method, the definition of the reward function, the inventory used by \graco{}, and the sizing of the components.

\subsection{Targeted Circuits, Testbenches, and Simulations}
\label{sec:tasks:subsec:circuit}

\noindent The circuits targeted in this paper are standard cells for digital systems: the \gls{inv} and the \gls{nand2}. These are circuits whose functionality and performance can be verified using a SPICE simulator. Consequently, the methods discussed in this paper can be applied to the design of any circuits, including both digital custom circuits and analog circuits. Table~\ref{tab:tasks} summarizes the settings for the two tasks that we consider.

The considered circuits are verified using a testbench that performs a transient analysis on the \gls{dut}. Input signal sources (e.g., input data, clock), supply source, and ground are instantiated in the testbench and connected to the available inputs of the \gls{dut} without relying on knowledge of its netlist (i.e., blackbox verification). The circuit's functionality is verified by comparing the sampled output voltage values with the expected values. Additionally, performance is assessed by measuring timing metrics such as rise/fall times and propagation delays. In this work, \ngspice{} is used as the circuit simulator.

Defining effective testbenches is crucial as they contribute to the reward which is the only metric that guides the synthesis process. The following guidelines are recommended when creating testbenches for \graco{}: 
\begin{itemize}
    \item \emph{Include all necessary measurements:} Ensures that the testbench captures all measurements required to verify the circuit and therefore to build the reward function. Missing measurements might prevent \graco{} from fully considering functional or performance requirements.
    \item \emph{Minimize testbenches and simulations:} Reducing the number of testbenches and simulations can lower the computational overhead.
    \item \emph{Set a timeout for non-converging simulations:} Including a timeout is recommended to handle cases where simulations fail to converge.
    \item \emph{Relax constraints for creativity:} Relaxing constraints can increase the design space exploration, potentially enabling \graco{} to discover creative circuit solutions. However, this involves a longer time taken for exploration.
\end{itemize}

\subsection{Reward function}
\label{sec:tasks:subsec:reward}

\begin{table*}[t]
\centering
\caption{Summary of considered design tasks. Sizes of topology design space are lower/upper bounds as discussed in Sec.~\ref{sec:expert_knowledge:subsec:hard_constraints}.}
\label{tab:tasks}
\resizebox{\linewidth}{!}{
\begin{tabular}{@{}lccccl@{}c@{}c@{}c@{}}
\toprule
\multirow{2}{*}{\textbf{Task}} & \multicolumn{3}{c}{\textbf{Available nets}} & \multirow{2}{*}{\textbf{Inventory}} & \multirow{2}{*}{\qquad\textbf{Sizing ranges}} & \multirow{2}{*}{\shortstack[c]{\textbf{Max sampling}\\\textbf{steps}}} & \multirow{2}{*}{\shortstack[c]{\textbf{Size of topology design space}\\\textbf{until standard gate}}} & \multirow{2}{*}{\shortstack[c]{\textbf{Size of topology design space}\\\textbf{until max. sampling steps}}}\\
\cmidrule(lr){2-4}
& Input & Output & Supply \\
\midrule
\raisebox{0.75\height}{Inverter} & \raisebox{0.75\height}{in} & \raisebox{0.75\height}{out} & \raisebox{0.75\height}{supply, gnd} & \shortstack[l]{\Verb|sky130_nfet_01v8|\\\Verb|sky130_pfet_01v8|} & \shortstack[c]{$w \in (0.36, 5), l \in (0.15, 5)$\\$w \in (0.42, 5), l \in (0.15, 5)$} & \raisebox{0.75\height}{6} & \shortstack[c]{$[1\cdot 10^3, 4\cdot 10^3]$ \\ ($1\times$NMOS, $1\times$PMOS)} &  \raisebox{0.75\height}{$[2\cdot 10^6, 1\cdot 10^9]$} \\
\\[-0.1cm]
\raisebox{0.75\height}{NAND2} & \raisebox{0.75\height}{in\textsubscript{1}, in\textsubscript{2}} & \raisebox{0.75\height}{out} & \raisebox{0.75\height}{supply, gnd} & \shortstack[l]{\Verb|sky130_nfet_01v8|\\\Verb|sky130_pfet_01v8|} & \shortstack[c]{$w \in (0.36, 5), l \in (0.15, 5)$\\$w \in (0.42, 5), l \in (0.15, 5)$} & \raisebox{0.75\height}{10} & \shortstack[c]{$[1\cdot 10^8, 8\cdot 10^{10}]$ \\ ($2\times$NMOS, $2\times$PMOS, $1\times$Internal net)}  &  \raisebox{0.75\height}{$[2\cdot 10^{12}, 4\cdot 10^{19}]$} \\
\bottomrule
\end{tabular}}
\vspace{-0.2cm}
\end{table*}

\noindent A reward function is required for the RL approach employed by \graco{}. Given the result of the SPICE simulation, it outputs a reward value bounded in $[-1,1]$, which \graco{} seeks to maximize. In this work, a reward of $1$ is assigned to a fully verified circuit based on the testbench results, at which point the sampling process can be terminated.

The reward function is constructed as the sum of multiple terms, referred to as \emph{subrewards}. It is defined as
\begin{align}
    r &= \frac{1}{L}\sum\nolimits_{l=1}^L s(r_{\text{sub},l}, r_{\text{min},l}),
    &
    r_{sub} &= 1 - \left[\frac{(m-m_\text{tar})}{m_\text{norm}}\right]^2,
    \label{eq:reward}
\end{align}
where $r$ is the reward, $L$ is the total number of subrewards, $r_\text{sub}$ is an individual subreward, $r_{min}$ is the minimum acceptable subreward, $s(r_\text{sub}, r_\text{min})$ is a saturation function, $m$ is the metric measured by the simulation, $m_\text{tar}$ is the target value, and $m_\text{norm}$ is a normalization factor.

Each subreward is calculated based on a simulation metric, such as a sampled voltage value or a timing value. The normalization factor ensures that the subreward is bounded within $[-1,1]$. The minimum subreward, $r_\text{min}$, is user-defined and represents the threshold for determining whether a subreward meets the circuit specifications. The saturation function $s(r_\text{sub}, r_\text{min})$ returns $1$ if $r_\text{sub} \geq r_\text{min}$, otherwise, it returns $r_\text{sub}$. This implements the incentive to maintain a good solution, as discussed in Sec.~\ref{sec:graco:subsec:discussion}. For our examples, circuit specifications determine $m$ in $r_\text{min}$, e.g., maximum delay of a standard cell, while $m_\text{tar}$ is the ideal target, e.g., a zero delay. However, it is worth noting that \graco{} could also be started without applying the saturation function to explore what design targets are achievable, which could then be used to determine $r_\text{min}$.

When all subrewards are maximized and equal to $1$, the overall reward also becomes $1$, indicating that the desired circuit has been found. At this point, the sampling process can be stopped.

\subsection{Inventory}
\label{sec:tasks:subsec:inventory}

\noindent As part of the graph generation, \graco{} can instantiate any component from the inventory. When targeting integrated circuits, the inventory can consist of the active and passive devices provided in a \gls{pdk}. In this work, all considered design tasks utilize NMOS and PMOS transistors from the SkyWater SKY130 \gls{pdk}~\cite{skywater130pdk}. However, \graco{} is not limited to synthesizing circuits exclusively from transistors. The inventory can also include subcircuits, which can significantly reduce the size of the design space, as \graco{} can reuse them, thereby decreasing the required number of sampling steps. For instance, storage cells can be designed from an inventory that includes only an inverter and a tristate inverter (i.e., an inverter whose output node can be set to high impedance based on an input clock signal). This capability of \graco{} not only enhances the efficiency of the design search but also facilitates the integration of existing \gls{ip} blocks as subcircuits, promoting the reuse of internal designs. When \gls{ip} blocks are used, their requirements can be considered by \graco{} to ensure their correct usage. 

\subsection{Sizing}
\label{sec:tasks:subsec:sizing}

\noindent \graco{} optimizes not only the circuit topology but also the sizing of circuit parameters. For tasks involving only NMOS and PMOS transistors from the SkyWater SKY130 \gls{pdk}, a range of minimum and maximum values is specified for the width and length of each transistor. Specifically, the minimum widths and lengths are defined by the \gls{pdk}, with minimum widths of 0.36\,$\mu$m for NMOS transistors and 0.42\,$\mu$m for PMOS transistors. The minimum transistor length for both is 0.15\,$\mu$m. For maximum values, 16\,$\mu$m is used for both widths and lengths. Within these ranges, \graco{} can sample parameter values according to a Gaussian distribution with inferred mean $\boldsymbol{\mu}_t$ and log-standard deviation $\log(\boldsymbol{\sigma}_t)$ from \eqref{eq:graco} to maximize the reward.

For \gls{ip} blocks, the sizing is often predefined but can also be sampled within specified ranges. When the sizing is fixed, \graco{} restricts its actions to instantiating and connecting the \gls{ip} block to surrounding elements in the circuit.

\section{Design Results}
\label{sec:results}

\begin{table*}[h!]
\centering
\caption{Effect of consistency checks on synthesis performance. Each sampling method was run three times, with results sorted by value. For convenience, we show the ``Optimality gap'' defined as $1 - $Best reward in the upper half of the table. Values below $1/15 = 0.067$ have the correct voltage output and do not exceed the maximum power constraints but differ in their timing behavior and are highlighted in \textcolor{blue}{blue}. We highlight the \textcolor{blue}{\textbf{best}} and \textcolor{blue}{\underline{second best}} circuit where the best one is on average $80\,$ps faster than the second best.}
\label{tab:nand2_results}
\resizebox{\linewidth}{!}{
\begin{tabular}{@{}lccccccccc@{}}
\toprule
\multirow{2}{*}{\textbf{Consistency Check}} & 
\multicolumn{3}{c}{\textbf{Random}} & 
\multicolumn{3}{c}{\textbf{\graco{} RLOO}} & 
\multicolumn{3}{c}{\textbf{\graco{} ES}} \\
\cmidrule(lr){2-4} \cmidrule(lr){5-7} \cmidrule(lr){8-10}
& Run $1$ & Run $2$ & Run $3$
& Run $1$ & Run $2$ & Run $3$
& Run $1$ & Run $2$ & Run $3$ \\
\midrule
\rowcol \multicolumn{10}{c}{\emph{Optimality gap $= 1 - $Best reward (smaller is better)}}\\
None                                                       & $1.33\cdot 10^{-1}$ & $1.33\cdot 10^{-1}$ & $1.33\cdot 10^{-1}$ & $2.25\cdot 10^{-1}$ & $3.14\cdot 10^{-1}$ & $3.62\cdot 10^{-1}$ & $8.13\cdot 10^{-2}$ & $1.33\cdot 10^{-1}$ & $1.33\cdot 10^{-1}$ \\
Connected in/out nets (during generation)                  & \textcolor{blue}{$4.08\cdot 10^{-4}$} & $6.96\cdot 10^{-2}$ & $1.33\cdot 10^{-1}$ & $1.38\cdot 10^{-1}$ & $2.18\cdot 10^{-1}$ & $4.00\cdot 10^{-1}$ & $6.74\cdot 10^{-2}$ & $7.51\cdot 10^{-2}$ & $1.33\cdot 10^{-1}$ \\
Paths between input and output nets (during generation)    & $6.96\cdot 10^{-2}$ & $1.33\cdot 10^{-1}$ & $1.33\cdot 10^{-1}$ & $2.39\cdot 10^{-1}$ & $2.40\cdot 10^{-1}$ & $2.45\cdot 10^{-1}$ & $6.74\cdot 10^{-2}$ & $1.33\cdot 10^{-1}$ & $1.33\cdot 10^{-1}$ \\
No floating nets (during generation)                       & $1.33\cdot 10^{-1}$ & $1.33\cdot 10^{-1}$ & $1.33\cdot 10^{-1}$ & $2.00\cdot 10^{-1}$ & $3.81\cdot 10^{-1}$ & $3.93\cdot 10^{-1}$ & $1.33\cdot 10^{-1}$ & $1.33\cdot 10^{-1}$ & $1.33\cdot 10^{-1}$ \\
No isolated subgraphs (during generation)                  & $1.33\cdot 10^{-1}$ & $1.33\cdot 10^{-1}$ & $1.34\cdot 10^{-1}$ & $2.39\cdot 10^{-1}$ & $3.80\cdot 10^{-1}$ & $3.81\cdot 10^{-1}$ & $1.31\cdot 10^{-1}$ & $1.33\cdot 10^{-1}$ & $1.33\cdot 10^{-1}$ \\
Connected in/out nets (after generation)                   & $7.05\cdot 10^{-2}$ & $1.33\cdot 10^{-1}$ & $1.34\cdot 10^{-1}$ & $1.60\cdot 10^{-1}$ & $3.81\cdot 10^{-1}$ & $3.96\cdot 10^{-1}$ & \textcolor{blue}{$7.73\cdot 10^{-4}$} & $1.30\cdot 10^{-1}$ & $1.33\cdot 10^{-1}$ \\
Paths between input and output nets (after generation)     & $1.33\cdot 10^{-1}$ & $1.33\cdot 10^{-1}$ & $1.33\cdot 10^{-1}$ & \underline{\textcolor{blue}{$1.34\cdot 10^{-4}$}} & $1.80\cdot 10^{-1}$ & $2.30\cdot 10^{-1}$ & $6.73\cdot 10^{-2}$ & $6.75\cdot 10^{-2}$ & $1.33\cdot 10^{-1}$ \\
No floating nets (after generation)                        & $7.39\cdot 10^{-2}$ & $1.33\cdot 10^{-1}$ & $1.33\cdot 10^{-1}$ & $1.38\cdot 10^{-1}$ & $3.67\cdot 10^{-1}$ & $3.81\cdot 10^{-1}$ & $6.71\cdot 10^{-2}$ & $1.33\cdot 10^{-1}$ & $1.34\cdot 10^{-1}$ \\
No isolated subgraphs (after generation)                   & $1.33\cdot 10^{-1}$ & $1.33\cdot 10^{-1}$ & $1.33\cdot 10^{-1}$ & $2.37\cdot 10^{-1}$ & $3.28\cdot 10^{-1}$ & $3.99\cdot 10^{-1}$ & $6.76\cdot 10^{-2}$ & $1.33\cdot 10^{-1}$ & $1.33\cdot 10^{-1}$ \\
Connected in/out nets (\graco{} input)                     & $-$ & $-$ & $-$ & $1.40\cdot 10^{-1}$ & $2.00\cdot 10^{-1}$ & $2.57\cdot 10^{-1}$ & $6.88\cdot 10^{-2}$ & $1.19\cdot 10^{-1}$ & $1.33\cdot 10^{-1}$ \\
Paths between input and output nets (\graco{} input)       & $-$ & $-$ & $-$ & $1.38\cdot 10^{-1}$ & $2.44\cdot 10^{-1}$ & $2.67\cdot 10^{-1}$ & $8.47\cdot 10^{-2}$ & $8.63\cdot 10^{-2}$ & $1.32\cdot 10^{-1}$ \\
No floating nets (\graco{} input)                          & $-$ & $-$ & $-$ & $1.34\cdot 10^{-1}$ & $1.53\cdot 10^{-1}$ & $4.00\cdot 10^{-1}$ & \textcolor{blue}{$\mathbf{6.66\cdot 10^{-5}}$} & $1.32\cdot 10^{-1}$ & $1.33\cdot 10^{-1}$ \\
No isolated subgraphs (\graco{} input)                     & $-$ & $-$ & $-$ & $1.33\cdot 10^{-1}$ & $2.46\cdot 10^{-1}$ & $3.30\cdot 10^{-1}$ & $6.74\cdot 10^{-2}$ & $1.33\cdot 10^{-1}$ & $1.33\cdot 10^{-1}$ \\
All (during generation)                                    & $6.77\cdot 10^{-2}$ & $1.33\cdot 10^{-1}$ & $1.33\cdot 10^{-1}$ & $2.67\cdot 10^{-1}$ & $3.23\cdot 10^{-1}$ & $3.26\cdot 10^{-1}$ & $6.79\cdot 10^{-2}$ & $8.77\cdot 10^{-2}$ & $1.33\cdot 10^{-1}$ \\
All (after generation)                                     & $1.33\cdot 10^{-1}$ & $1.33\cdot 10^{-1}$ & $1.34\cdot 10^{-1}$ & $1.83\cdot 10^{-1}$ & $2.27\cdot 10^{-1}$ & $3.61\cdot 10^{-1}$ & $6.78\cdot 10^{-2}$ & $7.15\cdot 10^{-2}$ & $1.33\cdot 10^{-1}$ \\
All (\graco{} input)                                       & $-$ & $-$ & $-$ & $2.01\cdot 10^{-1}$ & $2.67\cdot 10^{-1}$ & $3.14\cdot 10^{-1}$ & $1.30\cdot 10^{-1}$ & $1.32\cdot 10^{-1}$ & $1.33\cdot 10^{-1}$ \\
\midrule
\rowcol \multicolumn{10}{c}{\emph{Average train reward (higher is better; averaged over all training steps)}}\\
None                                                       & -0.430 & -0.430 & -0.428 & 0.567 & 0.594 & 0.625 & 0.247 & 0.255 & 0.284 \\
Connected in/out nets (during generation)                  & -0.436 & -0.435 & -0.435 & 0.523 & 0.592 & 0.648 & 0.241 & 0.265 & 0.275 \\
Paths between input and output nets (during generation)    & -0.437 & -0.435 & -0.434 & 0.580 & 0.586 & 0.675 & 0.244 & 0.266 & 0.269 \\
No floating nets (during generation)                       & -0.381 & -0.381 & -0.379 & 0.568 & 0.590 & 0.595 & 0.312 & 0.313 & 0.335 \\
No isolated subgraphs (during generation)                  & -0.430 & -0.430 & -0.430 & 0.592 & 0.597 & 0.654 & 0.246 & 0.252 & 0.297 \\
Connected in/out nets (after generation)                   & -0.228 & -0.228 & -0.226 & 0.589 & 0.597 & 0.648 & 0.242 & 0.266 & 0.276 \\
Paths between input and output nets (after generation)     & -0.222 & -0.222 & -0.221 & 0.581 & 0.646 & 0.840 & 0.250 & 0.257 & 0.270 \\
No floating nets (after generation)                        & -0.023 & -0.022 & -0.022 & 0.570 & 0.592 & 0.732 & 0.329 & 0.375 & 0.377 \\
No isolated subgraphs (after generation)                   & -0.426 & -0.425 & -0.425 & 0.551 & 0.596 & 0.597 & 0.250 & 0.256 & 0.279 \\
Connected in/out nets (\graco{} input)                     &    $-$ &    $-$ &    $-$ & 0.568 & 0.596 & 0.653 & 0.216 & 0.236 & 0.239 \\
Paths between input and output nets (\graco{} input)       &    $-$ &    $-$ &    $-$ & 0.592 & 0.631 & 0.735 & 0.219 & 0.219 & 0.246 \\
No floating nets (\graco{} input)                          &    $-$ &    $-$ &    $-$ & 0.527 & 0.575 & 0.596 & 0.208 & 0.229 & 0.233 \\
No isolated subgraphs (\graco{} input)                     &    $-$ &    $-$ &    $-$ & 0.594 & 0.597 & 0.668 & 0.214 & 0.223 & 0.239 \\
All (during generation)                                    & -0.437 & -0.434 & -0.433 & 0.579 & 0.580 & 0.643 & 0.259 & 0.283 & 0.288 \\
All (after generation)                                     & -0.132 & -0.130 & -0.130 & 0.591 & 0.592 & 0.601 & 0.278 & 0.287 & 0.294 \\
All (\graco{} input)                                       &    $-$ &    $-$ &    $-$ & 0.581 & 0.584 & 0.586 & 0.204 & 0.219 & 0.229 \\
\bottomrule
\end{tabular}}
\vspace{-0.25cm}
\end{table*}

\noindent As a baseline, we employ a random sampling approach in which all categorical distributions are assumed to be uniform, and the circuit parameter values are sampled uniformly from the specified ranges in the task description. This baseline represents a purely exploratory strategy, as it does not consider rewards from previous steps.

For RLOO and ES, we use the Adam optimizer with a linear learning rate warmup over five steps to a maximum of $10^{-3}$. A fixed decay schedule is applied after 50\% and 75\% of the training, with the learning rate reduced by a factor of $10$ at each step. The batch size is set to 256 to ensure sufficient diversity within each minibatch, which is crucial for obtaining a reliable leave-one-out baseline for the rewards. Training of \graco{} is conducted for a total of 1024 steps. For RLOO, an entropy regularization loss is added with a weight of $\lambda = 0.01$. For ES, the standard deviation for sampling perturbations $\boldsymbol{\epsilon}_k$ is set to $\sigma_\text{ES} = 0.05$.

\subsection{Inverter}
\label{sec:results:subsec:inv}

\begin{figure}[t]
    \vspace{-0.1cm}
    \centering
    \includegraphics[width=0.49\linewidth,trim=5 10 5 10,clip]{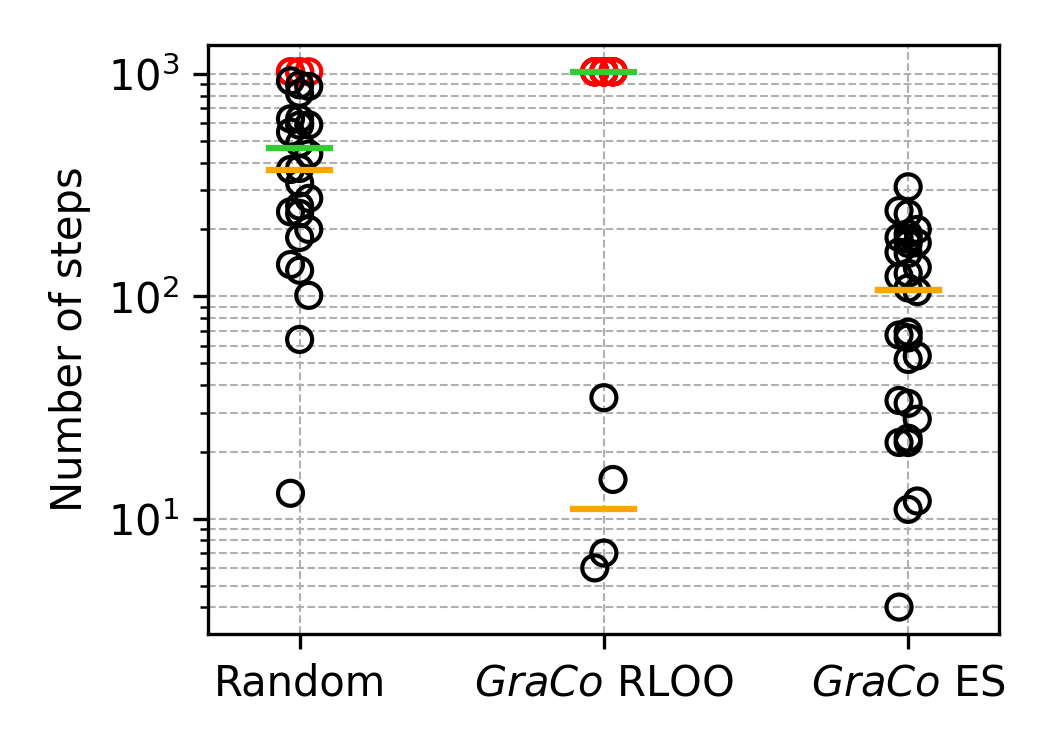}
    \hfill
    \includegraphics[width=0.49\linewidth,trim=5 10 5 10,clip]{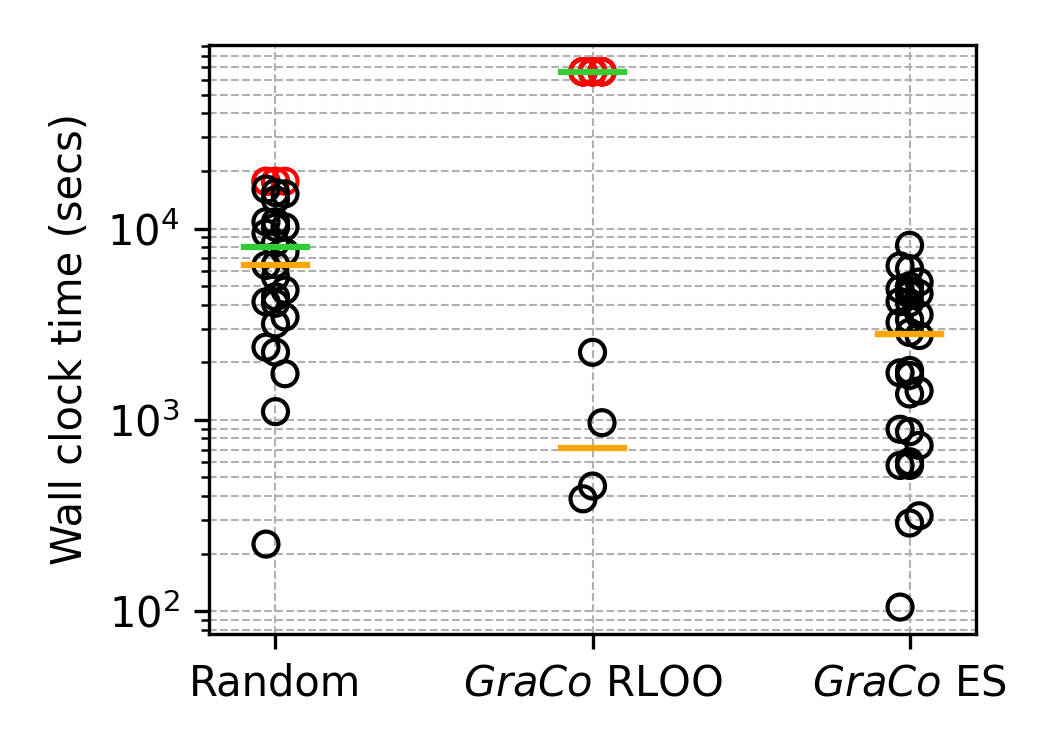}
    \caption{Distribution of sampling steps and wall clock time to find the standard inverter cell (reward = $1$). Each method was run $30$ times. Red circles mark failures by the random baseline or RLOO. Orange and green lines show median values for successful runs and all runs (including failures which iterated until the limit of $1024$ steps).}
    \label{fig:inv_sampling_steps}
    \vspace{-0.25cm}
\end{figure}

\noindent As a first design task, we address the synthesis of an inverter circuit. Using the task setup and reward definition from Sec.~\ref{sec:tasks}, we compare the efficiency of discovering the standard inverter cell across the three sampling methods. The results, shown in Fig.~\ref{fig:inv_sampling_steps}, indicate that ES reliably finds a verified inverter in all runs. In particular, ES quickly identifies the standard cell by effectively utilizing knowledge from previous steps. Compared to the random baseline, it requires approximately $5\times$ fewer steps and $3\times$ less wall clock time. On the other hand, RLOO suffers from insufficient exploration and focuses too heavily on exploitation, resulting often in solutions that fail to achieve the expected maximum reward of 1.0. However, there are instances where RLOO discovers the correct circuit very early in the synthesis process, making it in these cases much faster than both the random baseline and ES.

In summary, ES emerges as the best method, as it explores the design space more effectively. ES not only provides better sampling efficiency ($5\times$ less sampling steps than random), but also avoids collapsing too quickly to a sub-optimal solution, a known limitation of RLOO. This issue is evident in the worst-case samples in Fig.~\ref{fig:inv_sampling_steps}, where RLOO fails to find the standard cell and iterates until the maximum limit of $1024$ steps.

\subsection{NAND2}
\label{sec:results:subsec:nand2}

\noindent For the NAND2 design problem, we systematically compare the effects of all available consistency checks from Sec.~\ref{sec:expert_knowledge:subsec:checks} on the three sampling approaches. In this example, we do not employ a saturation function $s(r_\text{sub}, r_\text{min})$ for the measured timings, as the objective was to explore the fastest NAND2 that \graco{} can synthesize. Table~\ref{tab:nand2_results} presents the results of three runs, along with their averages.

First, examining the random baseline, we observe that applying consistency checks improves the average train rewards as expected. Notably, ensuring that there are no floating nets after generation and restarting the generation process if one fails (with a maximum of $10$ trials) leads to an average reward of $-0.02$ compared to $-0.43$ without consistency checks. This enhancement improves sampling efficiency by focusing only on potentially functional circuits. However, as the random baseline does not learn from previous rewards, the improved average reward does not directly translate into better performance in terms of the best reward.

For \graco{} RLOO, we observe a relatively high average train reward in the bottom part of Table~\ref{tab:nand2_results}, indicating its ability to reliably generate circuits with an average reward around $0.6$. However, it tends to converge too quickly, and despite the use of entropy regularization, it often does not achieve a high best reward. Consistency checks are expected to improve sampling efficiency. Our results indicate that they are often beneficial for achieving the best reward ($10$ out of $15$ times), although the randomness of the search can occasionally obscure this advantage.
Interestingly, we can observe that on average using the checks as additional input is the best choice, which we link to the fact that using the consistency checks during generation alters the sampling process which yields off-policy data that is more difficult to learn from using RLOO as is well known~\cite{munos2016safe}. Traditionally, importance sampling is used for REINFORCE to deal with this problem~\cite{chen2019top} which we plan to further investigate.

In contrast, \graco{} ES demonstrates slower learning, resulting in a lower average reward. However, its greater exploration of the design space allows it to find two of the four circuits with a reward exceeding $0.999$ which achieve the correct voltage output and do not exceed the maximum power constraints and only differ in their timing behavior. When comparing the results with and without consistency checks, we find that using consistency checks is beneficial $12$ out of $15$ times. Especially, during or after the generation process improves the performance: In both cases, using all checks together yields an average best reward of $0.90$ vs. $0.88$ for the baseline of not using any consistency check. However, utilizing all consistency checks as additional input features has not yet been effectively exploited by \graco{} if we use ES as learning strategy ($0.86$).

\begin{figure}
    \vspace{-0.1cm}
    \centering
    \includegraphics[width=0.34\linewidth,trim=15 10 10 5,clip]{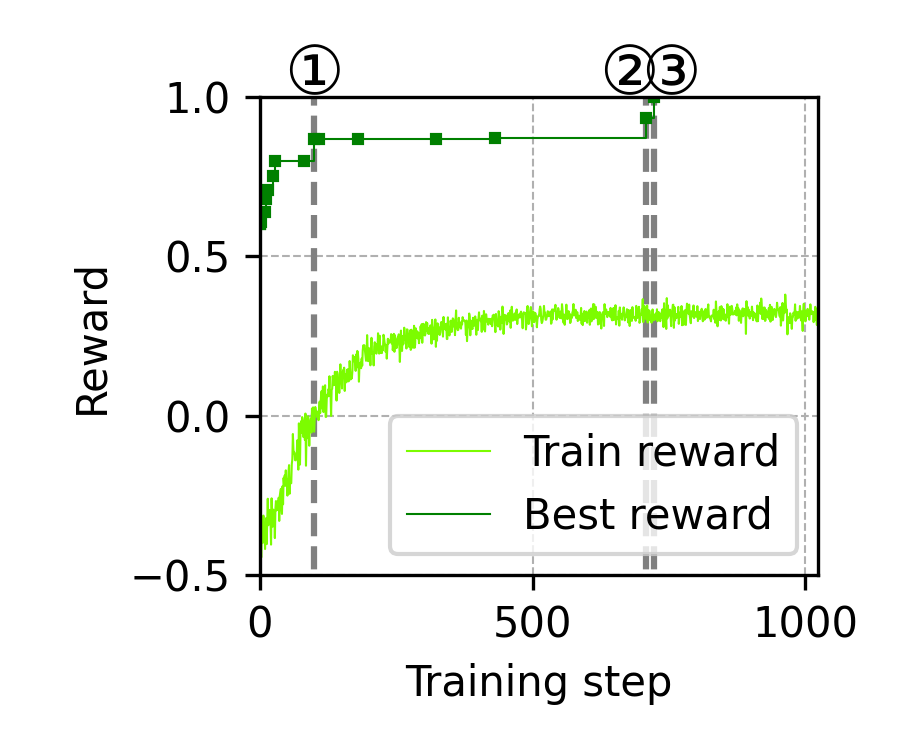}
    \hfill
    \includegraphics[width=0.64\linewidth,trim=5 -15 0 0,clip]{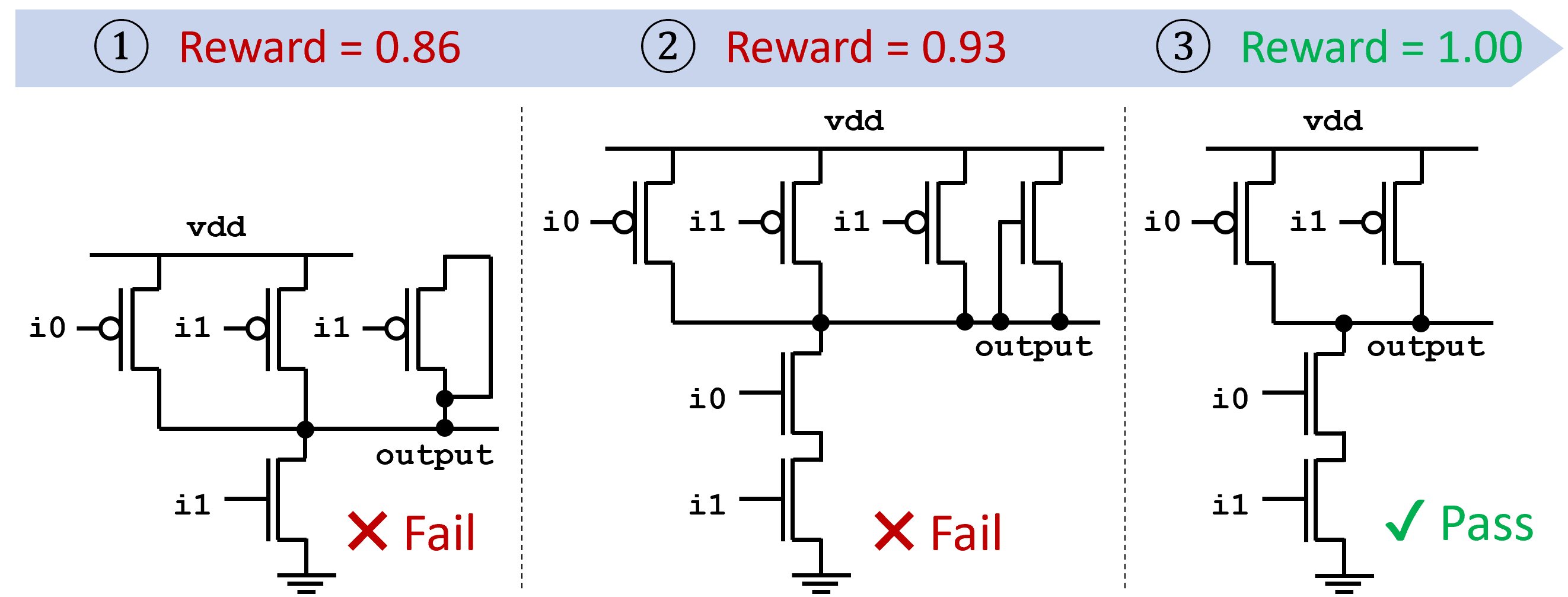}
    \caption{Evolution of generated NAND2 circuits for \graco{} ES. The left figure shows training reward per step and the best reward so far. The right figure illustrates circuit progression. Only the NAND2 circuit with reward $1$ passes the simulation-based verification.}
    \label{fig:nand2_progression}
    \vspace{-0.2cm}
\end{figure}

In summary, \graco{} ES can find a solution that is, on average, $30\%$ faster than the solution found by \graco{} RLOO and $2.5\times$ faster than the solution found by the random baseline. Finally, Fig.~\ref{fig:nand2_progression} illustrates the progression of the generated circuits. We observe that \graco{} ES progressively understands the behavior of the components, ultimately synthesizing the expected NAND2 gate.

\section{Conclusions and Future Work}
\label{sec:conclusions}

\noindent We introduced \graco{}, a method that uses reinforcement learning to construct integrated circuits step by step by building graphs that can be converted into netlists and verified with SPICE simulations. Our results demonstrate that implementing consistency checks often enhances the sampling efficiency of \graco{}. Additionally, we showed its capability to incorporate design knowledge effectively, enabling it to outperform a random sampling approach in designing circuits for various tasks. Compared to a random baseline, \graco{} achieves the design of an inverter with $5\times$ fewer sampling steps and successfully synthesizes a NAND2 gate that operates $2.5\times$ faster.

We propose the following directions to extend \graco{}: Firstly, we want to develop a mechanism to combine circuits with contrasting performance characteristics (e.g., rise vs. fall time) from different training runs. This idea, inspired by the recombination step in genetic algorithms, would allow leveraging the strengths of circuits with diverse performance traits. Secondly, we plan to enable \graco{} to utilize off-policy data. This involves facilitating training with combined circuits and existing designs by using a replay buffer that contains top-performing circuits and well-known designs from the literature. Thirdly, we aim to improve sampling efficiency by employing a hash table to store evaluated circuits. This would allow quick reward lookups and trigger resampling for topology duplicates, thereby reducing redundancy and enhancing exploration. This enhancement would particularly benefit the RLOO variant of \graco{}. Finally, we intend to extend \graco{} to handle larger and more complex tasks, paving the way for broader applicability in challenging design scenarios.

\newpage

\IEEEtriggeratref{16}
\bibliographystyle{IEEEtran}
\bibliography{references}

\begin{thebibliography}{10}
\providecommand{\url}[1]{#1}
\csname url@samestyle\endcsname
\providecommand{\newblock}{\relax}
\providecommand{\bibinfo}[2]{#2}
\providecommand{\BIBentrySTDinterwordspacing}{\spaceskip=0pt\relax}
\providecommand{\BIBentryALTinterwordstretchfactor}{4}
\providecommand{\BIBentryALTinterwordspacing}{\spaceskip=\fontdimen2\font plus
\BIBentryALTinterwordstretchfactor\fontdimen3\font minus \fontdimen4\font\relax}
\providecommand{\BIBforeignlanguage}[2]{{%
\expandafter\ifx\csname l@#1\endcsname\relax
\typeout{** WARNING: IEEEtran.bst: No hyphenation pattern has been}%
\typeout{** loaded for the language `#1'. Using the pattern for}%
\typeout{** the default language instead.}%
\else
\language=\csname l@#1\endcsname
\fi
#2}}
\providecommand{\BIBdecl}{\relax}
\BIBdecl

\bibitem{sripramong2002invention}
T.~Sripramong and C.~Toumazou, ``The invention of cmos amplifiers using genetic programming and current-flow analysis,'' \emph{IEEE transactions on computer-aided design of integrated circuits and systems}, vol.~21, no.~11, pp. 1237--1252, 2002.

\bibitem{mitea2011topology}
O.~Mitea, M.~Meissner, and L.~Hedrich, ``Topology synthesis of analog circuits with yield optimization and evaluation using pareto fronts,'' in \emph{2011 IEEE/IFIP 19th International Conference on VLSI and System-on-Chip}.\hskip 1em plus 0.5em minus 0.4em\relax IEEE, 2011, pp. 78--81.

\bibitem{meissner2014feats}
M.~Meissner and L.~Hedrich, ``{FEATS}: Framework for explorative analog topology synthesis,'' \emph{IEEE Transactions on Computer-Aided Design of Integrated Circuits and Systems}, vol.~34, no.~2, pp. 213--226, 2014.

\bibitem{yang2017smart}
Y.~Yang, H.~Zhu, Z.~Bi, C.~Yan, D.~Zhou, Y.~Su, and X.~Zeng, ``Smart-{MSP}: A self-adaptive multiple starting point optimization approach for analog circuit synthesis,'' \emph{IEEE Transactions on Computer-Aided Design of Integrated Circuits and Systems}, vol.~37, no.~3, pp. 531--544, 2017.

\bibitem{rojec2019analog}
{\v{Z}}.~Rojec, {\'A}.~B{\H{u}}rmen, and I.~Fajfar, ``Analog circuit topology synthesis by means of evolutionary computation,'' \emph{Engineering Applications of Artificial Intelligence}, vol.~80, pp. 48--65, 2019.

\bibitem{huss2006analog}
S.~A. Huss, ``Analog circuit synthesis: A search for the holy grail?'' in \emph{2006 IEEE International Symposium on Circuits and Systems}.\hskip 1em plus 0.5em minus 0.4em\relax IEEE, 2006, pp. 4--pp.

\bibitem{settaluri2020autockt}
K.~Settaluri, A.~Haj-Ali, Q.~Huang, K.~Hakhamaneshi, and B.~Nikolic, ``Autockt: Deep reinforcement learning of analog circuit designs,'' in \emph{2020 Design, Automation \& Test in Europe Conference \& Exhibition (DATE)}.\hskip 1em plus 0.5em minus 0.4em\relax IEEE, 2020, pp. 490--495.

\bibitem{zhao2020automated}
Z.~Zhao and L.~Zhang, ``An automated topology synthesis framework for analog integrated circuits,'' \emph{IEEE Transactions on Computer-Aided Design of Integrated Circuits and Systems}, vol.~39, no.~12, pp. 4325--4337, 2020.

\bibitem{fan2021specification}
S.~Fan, N.~Cao, S.~Zhang, J.~Li, X.~Guo, and X.~Zhang, ``From specification to topology: Automatic power converter design via reinforcement learning,'' in \emph{2021 IEEE/ACM International Conference On Computer Aided Design (ICCAD)}.\hskip 1em plus 0.5em minus 0.4em\relax IEEE, 2021, pp. 1--9.

\bibitem{budak2021dnn}
A.~F. Budak, P.~Bhansali, B.~Liu, N.~Sun, D.~Z. Pan, and C.~V. Kashyap, ``Dnn-opt: An {RL} inspired optimization for analog circuit sizing using deep neural networks,'' in \emph{2021 58th ACM/IEEE Design Automation Conference (DAC)}.\hskip 1em plus 0.5em minus 0.4em\relax IEEE, 2021, pp. 1219--1224.

\bibitem{zhao2022analog}
Z.~Zhao and L.~Zhang, ``Analog integrated circuit topology synthesis with deep reinforcement learning,'' \emph{IEEE Transactions on Computer-Aided Design of Integrated Circuits and Systems}, vol.~41, no.~12, pp. 5138--5151, 2022.

\bibitem{chen2023total}
Z.~Chen, S.~Meng, F.~Yang, L.~Shang, and X.~Zeng, ``Total: Topology optimization of operational amplifier via reinforcement learning,'' in \emph{2023 24th International Symposium on Quality Electronic Design (ISQED)}.\hskip 1em plus 0.5em minus 0.4em\relax IEEE, 2023, pp. 1--8.

\bibitem{lu2023automatic}
J.~Lu, L.~Lei, J.~Huang, F.~Yang, L.~Shang, and X.~Zeng, ``Automatic op-amp generation from specification to layout,'' \emph{IEEE Transactions on Computer-Aided Design of Integrated Circuits and Systems}, 2023.

\bibitem{dong2023cktgnn}
Z.~Dong, W.~Cao, M.~Zhang, D.~Tao, Y.~Chen, and X.~Zhang, ``Cktgnn: Circuit graph neural network for electronic design automation,'' \emph{arXiv preprint arXiv:2308.16406}, 2023.

\bibitem{budak2023apostle}
A.~F. Budak, D.~Smart, B.~Swahn, and D.~Z. Pan, ``Apostle: Asynchronously parallel optimization for sizing analog transistors using dnn learning,'' in \emph{Proceedings of the 28th Asia and South Pacific Design Automation Conference}, 2023, pp. 70--75.

\bibitem{zhao2024automated}
Z.~Zhao, J.~Liu, W.~Zhao, and L.~Zhang, ``Automated topology synthesis of analog integrated circuits with frequency compensation,'' \emph{IEEE Transactions on Computer-Aided Design of Integrated Circuits and Systems}, 2024.

\bibitem{aigner1999characterization}
M.~Aigner, ``A characterization of the bell numbers,'' \emph{Discrete mathematics}, vol. 205, no. 1-3, pp. 207--210, 1999.

\bibitem{graeb2001sizing}
H.~Graeb, S.~Zizala, J.~Eckmueller, and K.~Antreich, ``The sizing rules method for analog integrated circuit design,'' in \emph{IEEE/ACM International Conference on Computer Aided Design. ICCAD 2001. IEEE/ACM Digest of Technical Papers (Cat. No. 01CH37281)}.\hskip 1em plus 0.5em minus 0.4em\relax IEEE, 2001, pp. 343--349.

\bibitem{pretl2021fifty}
H.~Pretl and M.~Eberlein, ``Fifty nifty variations of two-transistor circuits: A tribute to the versatility of mosfets,'' \emph{IEEE Solid-State Circuits Magazine}, vol.~13, no.~3, pp. 38--46, 2021.

\bibitem{ohlrich1993subgemini}
M.~Ohlrich, C.~Ebeling, E.~Ginting, and L.~Sather, ``Subgemini: Identifying subcircuits using a fast subgraph isomorphism algorithm,'' in \emph{Proceedings of the 30th International Design Automation Conference}, 1993, pp. 31--37.

\bibitem{kunal2020gana}
K.~Kunal, T.~Dhar, M.~Madhusudan, J.~Poojary, A.~Sharma, W.~Xu, S.~M. Burns, J.~Hu, R.~Harjani, and S.~S. Sapatnekar, ``Gana: Graph convolutional network based automated netlist annotation for analog circuits,'' in \emph{2020 Design, Automation \& Test in Europe Conference \& Exhibition (DATE)}.\hskip 1em plus 0.5em minus 0.4em\relax IEEE, 2020, pp. 55--60.

\bibitem{ren2020paragraph}
H.~Ren, G.~F. Kokai, W.~J. Turner, and T.-S. Ku, ``Paragraph: Layout parasitics and device parameter prediction using graph neural networks,'' in \emph{2020 57th ACM/IEEE Design Automation Conference (DAC)}.\hskip 1em plus 0.5em minus 0.4em\relax IEEE, 2020, pp. 1--6.

\bibitem{hakhamaneshi2022pretraining}
K.~Hakhamaneshi, M.~Nassar, M.~Phielipp, P.~Abbeel, and V.~Stojanovic, ``Pretraining graph neural networks for few-shot analog circuit modeling and design,'' \emph{IEEE Transactions on Computer-Aided Design of Integrated Circuits and Systems}, vol.~42, no.~7, pp. 2163--2173, 2022.

\bibitem{li2019deepgcns}
G.~Li, M.~Muller, A.~Thabet, and B.~Ghanem, ``Deepgcns: Can gcns go as deep as cnns?'' in \emph{Proceedings of the IEEE/CVF international conference on computer vision}, 2019, pp. 9267--9276.

\bibitem{li2020deepergcn}
G.~Li, C.~Xiong, A.~Thabet, and B.~Ghanem, ``Deepergcn: All you need to train deeper gcns,'' \emph{arXiv preprint arXiv:2006.07739}, 2020.

\bibitem{zhang2019graph}
S.~Zhang, H.~Tong, J.~Xu, and R.~Maciejewski, ``Graph convolutional networks: a comprehensive review,'' \emph{Computational Social Networks}, vol.~6, no.~1, pp. 1--23, 2019.

\bibitem{williams1992simple}
R.~J. Williams, ``Simple statistical gradient-following algorithms for connectionist reinforcement learning,'' \emph{Machine learning}, vol.~8, pp. 229--256, 1992.

\bibitem{kool2019buy}
W.~Kool, H.~van Hoof, and M.~Welling, ``Buy 4 reinforce samples, get a baseline for free!'' in \emph{Proceedings of the ICLR 2019 Workshop: Deep RL Meets Structured Prediction}, 2019.

\bibitem{salimans2017evolution}
T.~Salimans, J.~Ho, X.~Chen, S.~Sidor, and I.~Sutskever, ``Evolution strategies as a scalable alternative to reinforcement learning,'' \emph{arXiv preprint arXiv:1703.03864}, 2017.

\bibitem{geweke1988antithetic}
J.~Geweke, ``Antithetic acceleration of monte carlo integration in bayesian inference,'' \emph{Journal of Econometrics}, vol.~38, no. 1-2, pp. 73--89, 1988.

\bibitem{skywater130pdk}
\BIBentryALTinterwordspacing
{Google and SkyWater Technology Foundry}, ``Skywater 130nm {PDK},'' 2020. [Online]. Available: \url{https://github.com/google/skywater-pdk}
\BIBentrySTDinterwordspacing

\bibitem{munos2016safe}
R.~Munos, T.~Stepleton, A.~Harutyunyan, and M.~Bellemare, ``Safe and efficient off-policy reinforcement learning,'' \emph{Advances in neural information processing systems}, vol.~29, 2016.

\bibitem{chen2019top}
M.~Chen, A.~Beutel, P.~Covington, S.~Jain, F.~Belletti, and E.~H. Chi, ``Top-k off-policy correction for a reinforce recommender system,'' in \emph{Proceedings of the Twelfth ACM International Conference on Web Search and Data Mining}, 2019, pp. 456--464.

\end{thebibliography}
\end{document}